\documentclass[times]{oupau}

\usepackage{graphicx}
\usepackage{enumitem}
\usepackage{tikz}
\usepackage{subcaption}
\usepackage{pifont}
\usepackage{hyperref}

\newcommand{\cmark}{\ding{51}}%
\newcommand{\xmark}{\ding{55}}%

\begin{document}
\title{TEDNet: Twin Encoder Decoder Neural Network for 2D Camera and LiDAR Road Detection}

\author{
Martín Bayón-Gutiérrez\affil{a}, 
María Teresa García-Ordás\affil{a},
Héctor Alaiz Moretón\affil{a},
Jose Aveleira-Mata\affil{a},
Sergio Rubio Martín\affil{b} and
José Alberto Benítez-Andrades\affil{b}
}

\address{
\affilnum{a}SECOMUCI Research Group, Department of Electric, Systems and Automatics Engineering, Universidad de León, Spain\\
\affilnum{b}SALBIS Research Group, Department of Electric, Systems and Automatics Engineering, Universidad de León, Spain\\
}

\corraddr{martin.bayon@unileon.es}

\begin{abstract}
Robust road surface estimation is required for autonomous ground vehicles to navigate safely. Despite it becoming one of the main targets for autonomous mobility researchers in recent years, it is still an open problem in which cameras and LiDAR sensors have demonstrated to be adequate to predict the position, size and shape of the road a vehicle is driving on in different environments. In this work, a novel Convolutional Neural Network model is proposed for the accurate estimation of the roadway surface. Furthermore, an ablation study has been conducted to investigate how different encoding strategies affect model performance, testing 6 slightly different neural network architectures. Our model is based on the use of a Twin Encoder-Decoder Neural Network (TEDNet) for independent camera and LiDAR feature extraction, and has been trained and evaluated on the Kitti-Road dataset. Bird's Eye View projections of the camera and LiDAR data are used in this model to perform semantic segmentation on whether each pixel belongs to the road surface. The proposed method performs among other state-of-the-art methods and operates at the same frame-rate as the LiDAR and cameras, so it is adequate for its use in real-time applications.
\end{abstract}


\received{\today}
\maketitle

\section{Introduction}
\label{sec:Introduction}
Autonomous driving research has suffered from an enormous increase in interest and research in the last few years, as society has paid more attention to advances in technologies that allow vehicles to perceive the surrounding environment and act in consequence \cite{Martinez2019}. Extensive use of Artificial Intelligence in combination with more traditional techniques helps solve key problems for autonomous vehicles such as object detection, pose estimation or traffic management. However, there are no hard rules on how to approach any of these topics, and usually, different groups select diverse approaches for their own autonomy platform.

One of the most important topics to be addressed to safely navigate in an unknown environment is to adequately estimate the position, size and shape of the road surface over which the vehicle must drive, a problem that is known as Road Detection or Road Estimation \cite{Aharon2014}. In this context, it is of interest to segment the roadway surface and estimate the free space in the environment prior to path planning the route for an autonomous vehicle \cite{Kulkarni2015}. In the case of Autonomous Ground Vehicles (AGVs), the path planning process is typically performed in a 2D space, also known as a Bird's Eye View (BEV) scene. Pretending the environment to be a flat surface over which vehicles move allows for avoiding system complexity in a context in which vertical elevation does not play an important role. In this scenario, a 2D BEV estimation of the road surface is of interest, as it that would align with the next steps in the autonomy pipeline.

In our previous work \cite{Bayón-Gutiérrez2022}, we presented a deep learning model for the road detection problem that made use of both LiDAR and camera sensors derived from U-net and VGG-16 models. In this work, a robust Road Estimation system for the Bird's Eye View perspective is presented. The system is based on the use of camera and LiDAR sensors and benefits from a novel Convolutional Neural Network (CNN) architecture to create a 2D confidence map of the road surface. The neural model has been trained on the Kitti-Road dataset, and an ablation study has been conducted to investigate how different parts of the CNN may be improved. A total of 6 different model architectures have been compared to investigate different data encoding techniques. Finally, authors present a CNN model that benefits for the use of a Twin Encoder-Decoder (TEDNet) to independently extract camera and LiDAR features that are useful for road semantic segmentation. Our model performs among other state-of-the-art methods on the Kitti-Road evaluation benchmark and opens discussion on how a Twin Encoder-Decoder architecture may be beneficial in the case of multi-sensor deep learning models.

The rest of this paper is structured as follows: In section \ref{sec:Related_Work} a revision of the road detection problem and some methods to face it is presented. Section \ref{sec:Material and Methods} introduces the Kitti-Road dataset, our data preprocessing technique, and all the CNN models that have been developed in this work. Results of our model are exposed in section \ref{sec:Experiments and Results}, along with a comparison with other state-of-the-art methods. Finally, in section \ref{sec:Conclusions} we present the conclusion of this work along with future research lines.

\section{Related work}
\label{sec:Related_Work}
Within the topic of Road Detection, the ultimate goal is to accurately detect those areas of the vehicle surroundings where it may be appropriate to drive \cite{Aharon2014}. The road area may be characterized by its size, shape, position, or even pavement color. Depending on the context of the particular problem to be solved, that area may be limited by the shape of the ego-lane in which the vehicle is driving on, or it may be extended to the complete paved area of the road, including other lanes and paved shoulders \cite{Ogden1997}.

Several approaches may be considered depending on the problem to be addressed, the sensor setup of the vehicle or the processing time requirements.

One of them is to detect road marking and use it to estimate the ego-lane in which the vehicle is at the time. In the case of multi-lane roads, it may be possible to additionally detect other lanes of the road, both for the direction of traffic and for the opposite direction \cite{Yenikaya2013}. This is usually the case for Advance Driver Assistance Systems (ADAS) that aim to prevent a human driver to unconsciously leave the lane \cite{Kumar2015}. In this case, front facing camera based systems perform well enough and can even be certified to comply with safety regulations \cite{Ponn2019}.
Nevertheless, this approach may not perform as expected in challenging situations like non-marked roads or if longitudinal cracks are wrongly detected as road markings \cite{Son2019}. 

Another strategy is to perform a semantic segmentation process, that aims to state whether the area that surrounds the vehicle corresponds to a road or not. This is commonly achieved by means of solving a single-class pixel-wise classification problem in which each pixel is assigned a binary label (i.e. Road or Not-Road) \cite{Garcia-Garcia2017} \cite{Hu2014}.
The semantic segmentation approach benefits from the possibility to not only detect the lane markings but to use image processing and deep learning techniques to detect the road according to its color, shape, appearance or even location.

Sensor setup and compute units to be used are also important when detecting the road surface. Authors in \cite{Xiao2016} achieved adequate results by the use of RGB cameras, which are inexpensive and easy to install on a test car. This has been the traditional approach to road detection like also discussed in \cite{Wang2021}. Camera-based systems perform well under most conditions but are heavily affected by light changes, and their use may not be possible in night-time situations, as discussed in \cite{Nayar1999}.

Another option is to use more advanced sensors such as LiDAR sensors. LiDAR sensors create a PointCloud of the environment by estimating the Time-of-Flight (ToF) of sensor-emitted laser beams, which are reflected in nearby objects. As LiDAR sensors generate their own laser pulses, they are not affected by light conditions, which is of importance if the road detection system is expected to work under low-light conditions or night-time. Works such as \cite{Hu2014,Caltagirone2017} have previously introduced the use of LiDAR sensors to perform semantic segmentation and road detection.

Authors in \cite{Bayón-Gutiérrez2022,Chen2019} exposed the benefit of performing sensor fusion techniques using both RGB images and LiDAR data for the road detection problem, leading to higher positions in road detection benchmarks, such as the Kitti-Road dataset leaderboard \cite{Fritsch2013}.

\section{Material and Methods}
\label{sec:Material and Methods}
Following a revision of the related work, the authors have developed a novel Deep Learning Road Detection Model. Our model is based on the use of an Encoder-Decoder architecture and employs both camera and LiDAR data to detect the paved surface of the road from a Bird's Eye View (BEV) perspective. An ablation study has also been conducted to compare slightly modified models and evaluate how the use of both sensors or just one of them affects the model performance.

\subsection{Kitti Dataset}
\label{sec:Kitti Dataset}
The Kitti Vision Benchmark Suite \cite{Geiger2013} is a well-known dataset for multiple types of Autonomous Vehicles related problems. Made public in 2012, it has become a reference in this field of study, as the Kitti suite is composed of several datasets and benchmarks including odometry estimation, 2D and 3D detection and tracking, semantic optical flow and road detection, among others.
For this work, authors have used the \emph{Kitti-Road dataset and evaluation benchmark} \cite{Fritsch2013} for training and evaluation of the developed model.

This dataset presents a number of scenes in several road environments (e.g. residential vs suburban or marked vs unmarked), for which the road surface and ego-lane have been annotated. The dataset contains 289 training scenes and 290 test scenes. It is important to note that, annotations for the test scenes are not publicly available to prevent training in this batch of scenes. Instead, the dataset authors provide an evaluation server on the dataset webpage. Authors may submit their estimations for the test scenes and get an evaluation score and entry on the dataset leaderboard.

The data was collected in 2011 by installing different sensors on the roof or a car, and each scene is composed of a pair of color and grayscale stereo images from forward-pointing cameras, 3D LiDAR data from a Velodyne HDL-64E sensor installed in the middle of the vehicle roof and GPS and IMU measurements. All sensors have been synchronized so the cameras are triggered when the LiDAR sensor is pointing forward and scenes in the dataset are sorted in different categories (i.e. Urban Unmarked (UU), Urban Marked (UM) and Urban Multiple Marked (UMM)). For this work, authors have opted to just use LiDAR data and the left color camera images, along with ground truth labels.

As indicated in the dataset paper, evaluation of the methods is performed in the BEV space. This space spans 20 meters laterally (-10m, 10m) and 40 meters longitudinally (6m, 46m) in front of the vehicle. A resolution of 0.005m/px leads to images of 800$\times$400 pixels.

\subsection{Data preprocessing}
\label{sec:Data Preprocessing}
Authors made use of the development tools available along the Kitti-Road dataset to adapt the left color camera images and Ground Truth labels to the BEV space. This operation needs to be done for each image before it is fed to the CNN.

The Velodyne LiDAR raw data is presented in the dataset as floating point binaries. Here, each individual point is represented by 3 values for the \emph{x}, \emph{y} and \emph{z} axes and an additional value for the intensity channel.
As the evaluation of the road detection is performed in the 2D BEV perspective, authors developed their own preprocessing tool to adapt the 3D LiDAR data to the BEV perspective as described in their previous work \cite{Bayón-Gutiérrez2022}. A similar approach for LiDAR preprocessing has also been used in \cite{Caltagirone2017, Yu2021}. This operation needs to be done for each LiDAR PointCloud before it is fed to the CNN.

Each LiDAR PointCloud is first clipped to the 20 x 40 meters Region of Interest (RoI) of the dataset, and then transformed into a 800$\times$400$\times$3 RGB image that corresponds to a vertical view of the scene.
For this, a grid of equal size of the image is created, and each point of the PointCloud is assigned to the corresponding cell according to its \emph{x} and \emph{y} values. Each of the LiDAR image channels are then created following these constraints: 

\begin{itemize}
    \item The Red channel of the image corresponds to whether each cell of the grid contains at least one point (255) or not (0).
    \item The Green channel of the image encodes the mean intensity of the points that belong to each cell of the grid. The value is normalized in the [0-255] range.
    \item The Blue channel of the image encodes the mean \emph{z} value of the points that belong to each cell of the grid. The value is truncated to the [-1.8m, -1.2m] range and then normalized in the [0-255] range. This prevents outliers to have a high influence in this channel.
\end{itemize}

The result of the LiDAR preprocessing is a 3-channel image of 800$\times$400 pixels that represents the 3D PointCloud in the same BEV perspective as the camera images.

\begin{figure}
    \centering
    \includegraphics[width=0.9\textwidth,,keepaspectratio]{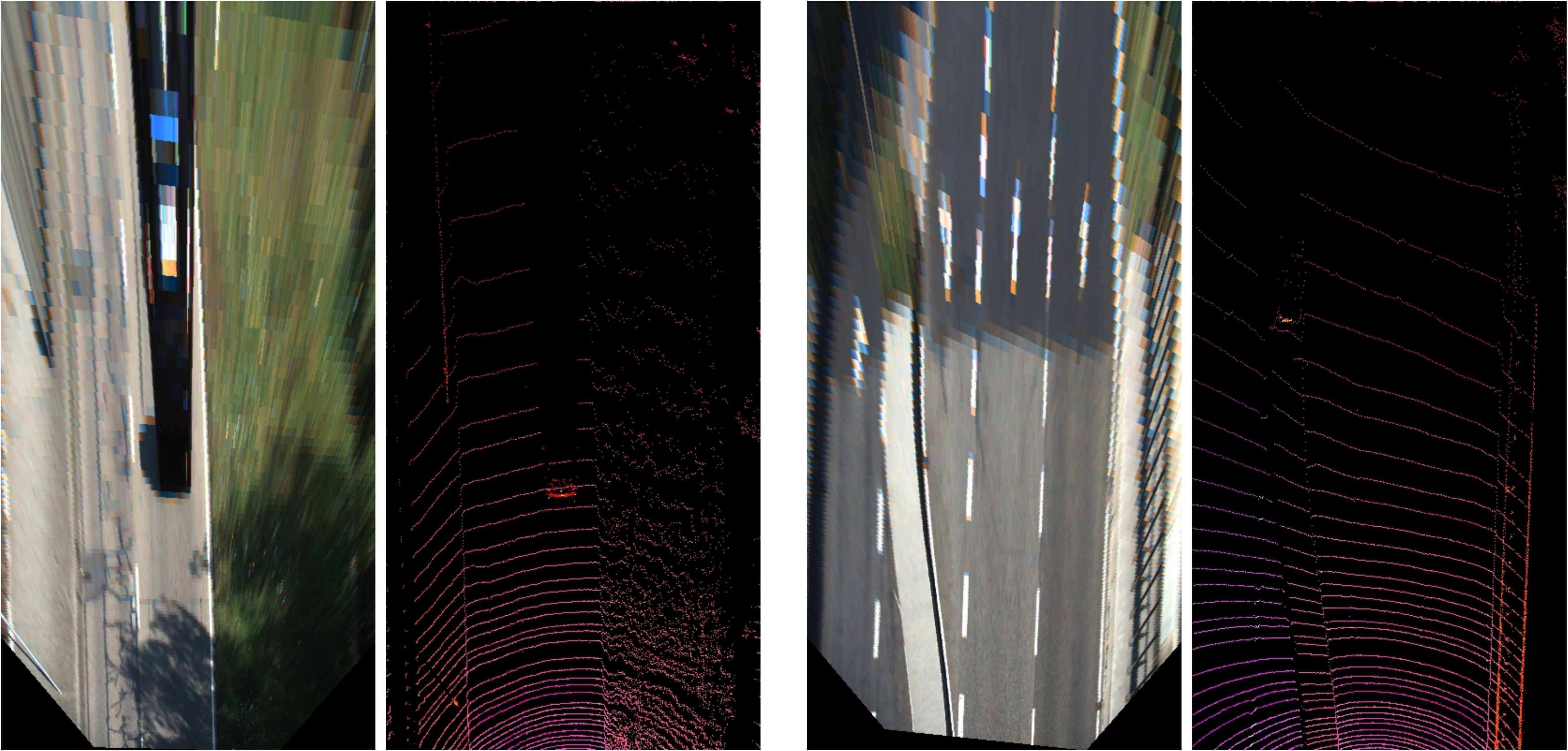}
    \caption{Preprocesed camera and LiDAR data}
    \label{fig:PreprocessedData}
\end{figure}

Depending on the CNN architecture needs, two different approaches can be applied:

\begin{itemize}
    \item As the camera and LiDAR images represent then the same spatial space, they can be combined to form a single 800$\times$400$\times$6 pixels image. Here, a 6-channel image is fed to the neural network model as a single data structure.
    \item Camera and LiDAR data can also be used as independent RGB images, so different inputs from the neural network may receive the camera or LiDAR data independently. An example of independent camera and LiDAR data is presented in Figure \ref{fig:PreprocessedData}
\end{itemize}

Both scenarios have been studied in this work, as indicated in section \ref{sec:Modifications to the base model}.

\subsection{Base Neural Network}
\label{sec:Base Neural Network}
Based on our previous work \cite{Bayón-Gutiérrez2022}, authors have developed a novel Convolutional Neural Network (CNN) following a symmetrical Encoder-Decoder architecture with two fully connected and dropout layers between them. The model aims to perform road semantic segmentation in the BEV space.

An Encoder-Decoder Neural Network is a type of Deep Learning model in which the first part of the model (Encoder) performs a feature extraction process that leads to a high-dimensional feature vector, this feature vector is then used by the second part of the model (Decoder) to generate a low-dimensional output similar to the input of the Neural Network. For semantic segmentation problems, an Encoder-Decoder architecture typically outputs a semantic map with the same shape of the input image.
 
In this work, the Encoder part of the network is composed of 4 Convolution blocks followed by Max-Pooling layers. In each of these blocks, the depth of the network grows from the original 6 channels input, to 16, 32, 64 and finally 128 channels, and the number of convolutional layers in each block varies between 2 and 3. The output of the Encoder is a 50x25x128 data structure.

The result of the Encoder is then fed to two 50x25x1024 dense layers with dropout regularization to reduce the probability of overfitting to the training dataset. 

The Decoder of the Neural Network is symmetrical with respect to the encoder. This means that 2D transposed convolutional layers, followed by a variable number of convolutional layers are used to reduce the depth of the data structures from 128, to 64, 32, 16 and finally a single channel.

In addition, the authors have implemented 3 skip connections from the end of convolutional blocks 2, 3 and 4 of the Encoder to the beginning of convolutional blocks 1, 2 and 3 of the Decoder. These skip connections are made possible because of the symmetrical aspect of the model, which ensures that the size of connected layers is the same, so the skip connection is implemented as the combination of the current layer with the result of a previous layer. Skip connections have proved to help reach model convergence in deep neural models and Encoder-Decoder models \cite{Wang2021,Ronneberger2015}.

For this model, the first of the two approaches for Input data presented in section \ref{sec:Data Preprocessing} is applied. 800$\times$400$\times$6 images combining camera and LiDAR data are used as input for the CNN, and a 800$\times$400$\times$1 image is generated as the output of the model.
This single-channel image corresponds to a confidence map for the road detection process, where each pixel is assigned a value between (0, 1) according to whether that pixel corresponds to a road in the BEV perspective or not.
This process is commonly known as semantic segmentation and, in our case, binary semantic segmentation, as the CNN model just aims to predict whether each pixel in the input image corresponds to a road surface or not.

The model has been implemented in Python, using the Keras Functional API and Figure \ref{fig:ModelsArchitecture} presents a representation of the model. Further details relating to training and evaluation of the model are presented in section \ref{sec:Training and evaluation}.

Once implemented and tested, authors have modified this base model to produce a total of 6 slightly different models, so that several implementations could be compared in the same dataset.

\subsection{Modifications to the base model}
\label{sec:Modifications to the base model}
An ablation study has been conducted to refine the model and to evaluate the performance of similar architectures trained on the same dataset.
As one of the aims of this work was to evaluate how the use of camera and LiDAR data may impact the results of the given network model, several data input strategies have been tested. In all of the cases, the output of the model is a 800$\times$400 single-channel confidence map representing road detection.

\begin{figure}
    \centering
    \includegraphics[width=0.9\textwidth,height=\textheight,keepaspectratio]{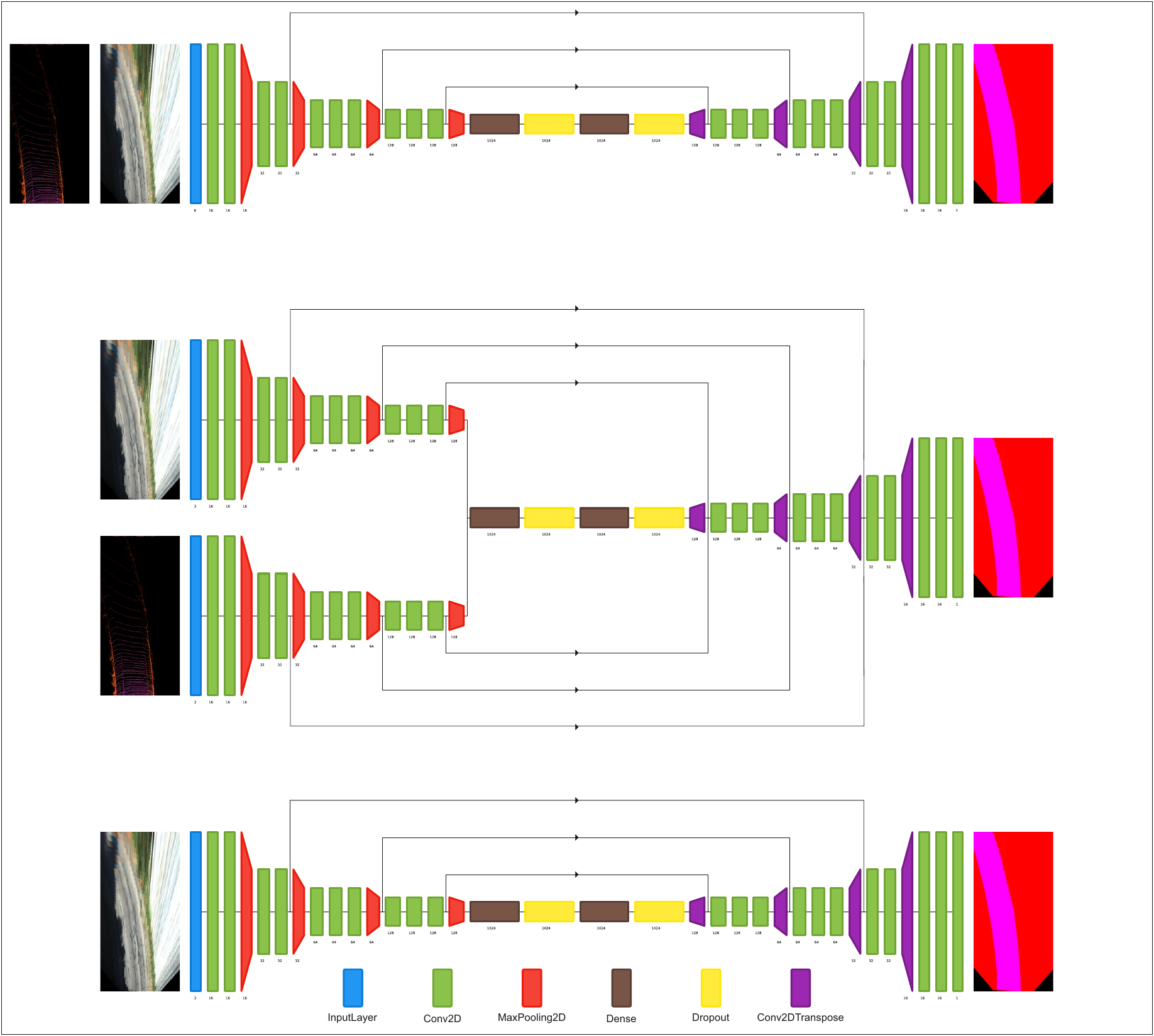}
    \caption{From top to bottom, CNN architecture for models A, C and E}
    \label{fig:ModelsArchitecture}
\end{figure}

\begin{itemize}
    \item \textbf{Model A:} Base model\\
    This architecture corresponds to the base model as presented in section \ref{sec:Base Neural Network}. The input of the model is a 800$\times$400$\times$6 image corresponding to the combined camera and LiDAR BEV images. Skip connections are present in this model. This model is represented in Figure \ref{fig:ModelsArchitecture}
    
    \item \textbf{Model B:} No Skip connections\\
    This architecture is similar to the base model (Model A), but the 3 skip connections have been removed from the network. This Model aims to investigate on how the skip connections affect the model performance.
    
    \item \textbf{Model C:} Independent twin encoders\\
    To investigate on different data ingestion strategies, authors apply in this model the second approach for input data presented in section \ref{sec:Data Preprocessing}. This implies that the input is composed by two 3-channel RGB images for the camera and LiDAR data.
    Authors have created a twin Encoder model in which the weights of the encoders are independent from each other (i.e. the Encoders are non-siamese). One of the Encoders receives the camera RGB image, while the other receives the LiDAR RGB representation.
    This Model aims to investigate on how the network may benefit from the independent treatment of camera and LiDAR images during the feature extraction process that takes place in the Encoder part of the network.
    Skip connections are applied from both of the encoders as in Model A. This model is represented in Figure \ref{fig:ModelsArchitecture}
    
    \item \textbf{Model D:} Independent twin encoders with no Skip connections\\
    This architecture is similar to the previous model (Model C), but the 3 skip connections have been removed from the network (as in Model B). This model aims to investigate on how the skip connections affect the model performance.
    
    \item \textbf{Model E:} Only camera data\\
    This model is derived from Model C, but the encoder for the LiDAR data has been removed. Just camera images are fed to the model with the aim to evaluate how performance is affected in case of LiDAR data is not present. This model is represented in Figure \ref{fig:ModelsArchitecture}
    
    \item \textbf{Model F:} Only LiDAR data\\
    This model is derived from Model C, but the encoder for the camera data has been removed. Just LiDAR images are fed to the model with the aim to evaluate how performance is affected in case of camera data is not present.
\end{itemize}

To make sure that all the models performance is evaluated in equal conditions, authors have investigated the  best configuration for the Model A, and then replicated this configuration for the other models, so results just depend on the architecture of the model and sensor setup utilized.

\subsection{Training and evaluation}
\label{sec:Training and evaluation}
As indicated previously, the authors have opted to use the Kitti-Road dataset for the training and evaluation of the models. This dataset just contains 289 annotated scenes, so training models is often challenging due to the small number of available samples.

In order to get the most of the dataset, authors have implemented several techniques to improve model performance.
\begin{itemize}  
    \item Grid-search:\\
    Hyper-parameters tuning is one of the most important part of the development of a deep learning model, as the same model can provide drastically different performance results with a bad hyper-parameters configuration with respect to an adequate one.
    Although some expertise and common good-enough values can be used to fine-tune the network hyper-parameters, the implementation of a grid search technique usually leads to better results.
    Grid-search allows for the testing of several configurations of hyper-parameters. For each configuration, the model will be trained and evaluated, so it is possible to determine which configuration provided the best results and assure that no other combination of the hyper-parameters that were tested could perform better.

    In this work, the grid-search included the following hyper-parameters: optimizer (Adam or SGD), train/validation-split (Between 0.1 and 0.5), loss function (Binary Cross-Entropy or Dice-Loss), Learning rate, Dropout rate and Data augmentation rate.

    Grid-search was applied to Model A to find the best combination of hyper-parameters. Those were later used to train all the models and obtain the results exposed in section \ref{sec:Experiments and Results}.

    \item Cross-validation:\\
    In cases such as the one presented in this work, in which the dataset used for training have a small number of samples, it is important to prevent overfitting to the training and validation datasets.
    To account for this problem, authors have implemented Cross-validation during the training and evaluation of all the models

    The grid search results showed that the best train/validation split for the model was to use 90\% of the dataset as the training dataset and the 10\% remaining scenes as the validation dataset.
    Taking this results into consideration, authors implemented a k-fold cross-validation technique with k=10, so for each fold of the cross-validation process, the validation dataset corresponds to a 10\% of the complete dataset.

    \item Data augmentation:\\
    Data augmentation techniques are applied to compensate for the small number of samples present in the Kitti dataset.
    As the different models that are evaluated in this work present different input data strategies, a simple data augmentation process has been implemented.

    First, a data augmentation rate is selected using the results from the grid-search. Then, a portion of the training dataset is arbitrary selected for the data augmentation process.
    As the grid-search preliminary results showed that data augmentation rate = 1 performed the best, data augmentation is applied to the entire training dataset.
    The scenes that are selected for the data augmentation process are duplicated, and one of the copies is flipped left-to-right.
    This simple data augmentation process allows the network to train in up to twice the number of scenes of the training dataset.
\end{itemize}

The authors included several metrics to evaluate the performance of the CNN.
We employ the same metrics that will be taken into account in the Kitti-Road benchmark: Maximum F1-score, Precision, Recall, False Positive Rate and False Negative Rate. In addition to these, we have also included BinaryIoU and we use this as our main metric to compare among the 6 models that were tested, as IoU, and particularly BinaryIoU, have demonstrated greater performance in semantic segmentation problems than other pixel-based metrics.

\section{Experiments and Results}
\label{sec:Experiments and Results}
A desktop PC with the following characteristics was employed for the development and training of the proposed method.

\begin{itemize}  
    \item Intel Core I9-9900K CPU @ 3,60 GHz 
    \item 48GB DDR4 RAM
    \item Nvidia RTX 2080 Ti 11GB GPU
    \item Ubuntu 22.04 LTS OS
    \item Python 2.7 / Python 3.9
    \item Tensorflow 2.8.0
\end{itemize}

\begin{figure}
    \centering
    \includegraphics[width=0.9\textwidth,height=\textheight,keepaspectratio]{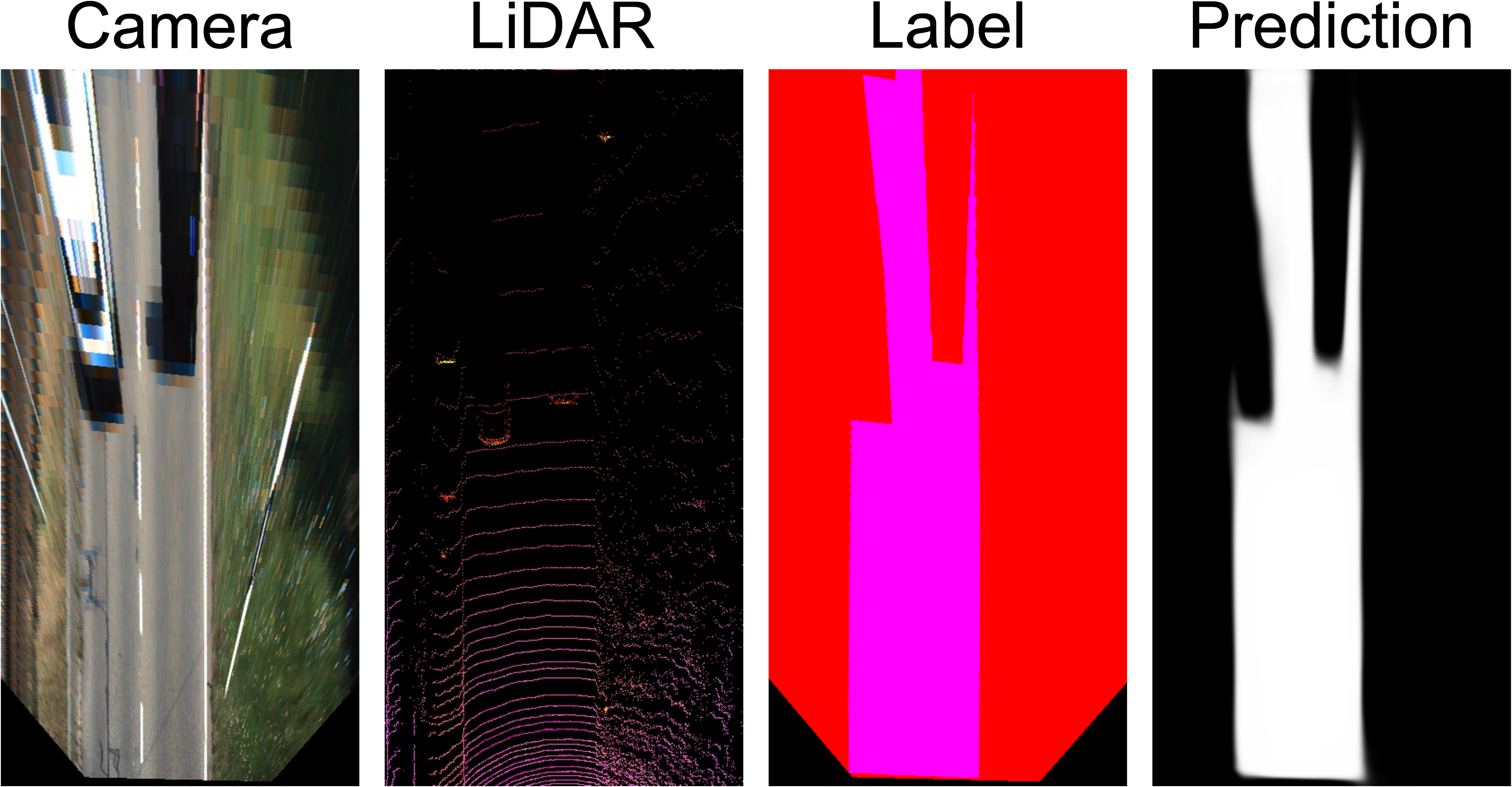}
    \caption{Model prediction and true label on one scene of the validation split}
    \label{fig:Examples}
\end{figure}

All models from section \ref{sec:Modifications to the base model} were trained using the same dataset and hyperparameters configuration. Furthermore, we use the same seed for all the models, so weight initialization and dataset shuffling are consistent. Figure \ref{fig:Examples} presents examples of the road detection process on the validation dataset for one of our models.

\subsection{Comparison among models}
\label{sec:Comparison among models}
As introduced in section \ref{sec:Training and evaluation}, we compare all the developed models according to the BinaryIoU metric. For this matter, we compute the validation BinaryIoU metric for each of the 10 folds of the cross-validation process, and compare the average value among all the models in Table \ref{tab:ModelsComparison}.

\begin{table}[]
    \centering
    \caption{Comparison among the different models}
    \label{tab:ModelsComparison}
    \begin{tabular}{c | c | c | c | c | c }
    \bf{Model}	& \bf{Camera}	& \bf{LiDAR}	& \bf{Combined input}	& \bf{Skip connections}	& \bf{BinaryIoU}        \\ \hline
    Model A		& \cmark		& \cmark		& \cmark				& \cmark				& 0.9065					\\
    Model B		& \cmark		& \cmark		& \cmark				& \xmark				& 0.9056					\\
    Model C		& \cmark		& \cmark		& \xmark				& \cmark				& 0.9118					\\
    Model D		& \cmark		& \cmark		& \xmark				& \xmark				& 0.9135					\\
    Model E		& \cmark		& \xmark		& \xmark				& \cmark				& 0.8859					\\
    Model F		& \xmark		& \cmark		& \xmark				& \cmark				& 0.8899					\\
    \end{tabular}
\end{table}

Model D yields the best results in our experiments, with an average BinaryIoU of 0.9135, followed closely by Model C, with an average BinaryIoU of 0.9118. These models correspond to the architecture with independent encoders for the LiDAR and camera data. Here, the model without skip connections performs slightly better.

Models A and B, those with a combined encoder, offer a similar average BinaryIoU between them, both over 0.90.
Models that just employ one of the sensors, Models E and F, performed the worse, with 0.8859 and 0.8899 respectively.

These results state that, there is a clear difference between the models performance, especially when big changes are made to the CNN architecture. The existence of the skip connections does not seem to affect the models results as much as the different Encoder architectures does.

The difference between models E and F and all others demonstrates that, for this use case, the use of both sensors is of importance, as single-sensor models perform worse than those who use both sensors in one way or another.

All the models have a similar inference time of around 0.091 seconds/image ($\approx$ 10fps), which is equivalent to the LiDAR and camera sampling rate, which means that the model can run in real-time.

\subsection{Kitti Dataset results}
Models C and D were used to predict the road on the testing dataset, as those are the one that offered the best results among the models that were tested.
The results for the road detection were uploaded to the Kitti-Road evaluation server, and the results from the external evaluation are presented in Table \ref{tab:KittiResultsTable}.

\begin{table}[]
    \caption{Results from the evaluation server for models C and D}
    \label{tab:KittiResultsTable}
    \centering
    \begin{tabular}{c | c | c | c | c | c | c}
    {\bf Benchmark}	& {\bf MaxF}	& {\bf AP}		& {\bf PRE}	& {\bf REC}	& {\bf FPR}	& {\bf FNR}	\\ \hline
    \multicolumn{7}{c}{Model C}\\ \hline
    UM\_ROAD		& 94.24 \%		& 92.43 \%	& 93.45 \%	& 95.04 \%	& 3.04 \%	& 4.96 \%	\\
    UMM\_ROAD		& 95.94 \%		& 95.31 \%	& 96.21 \%	& 95.67 \%	& 4.14 \%	& 4.33 \%	\\
    UU\_ROAD		& 92.78 \%		& 91.08 \%	& 91.86 \%	& 93.71 \%	& 2.71 \%	& 6.29 \%	\\
    URBAN\_ROAD		& 94.62 \%		& 93.05 \%	& 94.28 \%	& 94.96 \%	& 3.17 \%	& 5.04 \%	\\ \hline    
    \multicolumn{7}{c}{Model D}\\ \hline
    UM\_ROAD		& 93.19 \%	& 92.51 \%	& 92.42 \%	& 93.97 \%	& 3.51 \%	& 6.03 \%	\\
    UMM\_ROAD		& 94.65 \%	& 95.07 \%	& 94.39 \%	& 94.91 \%	& 6.20 \%	& 5.09 \%	\\
    UU\_ROAD		& 92.00 \%	& 90.57 \%	& 90.87 \%	& 93.16 \%	& 3.05 \%	& 6.84 \%	\\
    URBAN\_ROAD		& 93.53 \%	& 92.96 \%	& 92.87 \%	& 94.19 \%	& 3.98 \%	& 5.81 \%	\\
    \end{tabular}
\end{table}

Contrary to our preliminary tests, Model C performed better in the test dataset, for which ground truth labels were not available. Authors have submitted the results for Model C to the \href{http://www.cvlibs.net/datasets/kitti/eval\_road.php}{Kitti-Road benchmark leaderboard} under the acronym TEDNet.

Table \ref{tab:KittiLeaderboard} presents a comparison between our models and other state-of-the-art methods that employ both LiDAR and camera sensors to detect the road surface. 
Our method not only performs similarly in terms of pixel-wise evaluation but is also comparable in terms of running time.

\begin{table}[htp!]
    \centering
    \caption{Kitti-Road leaderboard for methods that use both camera and LiDAR data}
    \label{tab:KittiLeaderboard}
    \begin{tabular}{c | c | c | c | c | c | c | c}
    {\bf Method} & {\bf Year} & {\bf MaxF} & {\bf AP} & {\bf Time} & {\bf Cite} \\ \hline
    PLARD			& 2019 & 97.03 \% & 94.03 \% & 0.16s & \cite{Chen2019}	\\
    LRDNet+			& 2023 & 96.95 \% & 92.22 \% & 0.10s & \cite{Khan2022}\\
    CLCFNet			& 2021 & 96.38 \% & 90.85 \% & 0.02s & \cite{Gu2021}\\
    LidCamNet		& 2018 & 96.03 \% & 93.93 \% & 0.15s & \cite{Caltagirone2018}\\
    LC-CRF			& 2019 & 95.68 \% & 88.34 \% & 0.18s & \cite{Gu2019Road}\\
    BJN             & 2021 & 94.89 \% & 90.63 \% & 0.02s & \cite{Yu2021}\\
    \hline
    {\bf TEDNet (ours)} & 2023 & 94.62 \% & 93.05 \% & 0.09s & \\
    \hline
    CLRD			& 2022 & 94.20 \% & 92.66 \% & 0.05s & \cite{Bayón-Gutiérrez2022}\\
    HID-LS			& 2019 & 93.11 \% & 87.33 \% & 0.25s & \cite{Gu2019Histograms}\\
    HybridCRF		& 2018 & 90.81 \% & 86.01 \% & 1.50s & \cite{Xiao2018}\\
    MixedCRF		& 2017 & 90.59 \% & 84.24 \% & 6.00s & \cite{Han2017}\\
    FusedCRF		& 2015 & 88.25 \% & 79.24 \% & 2.00s & \cite{Xiao2015}\\
    RES3D-Velo		& 2014 & 86.58 \% & 78.34 \% & 0.36s & \cite{Shinzato2014}\\
    \end{tabular}
\end{table}

\begin{figure}
    \centering
    \includegraphics[width=0.9\textwidth,height=\textheight,keepaspectratio]{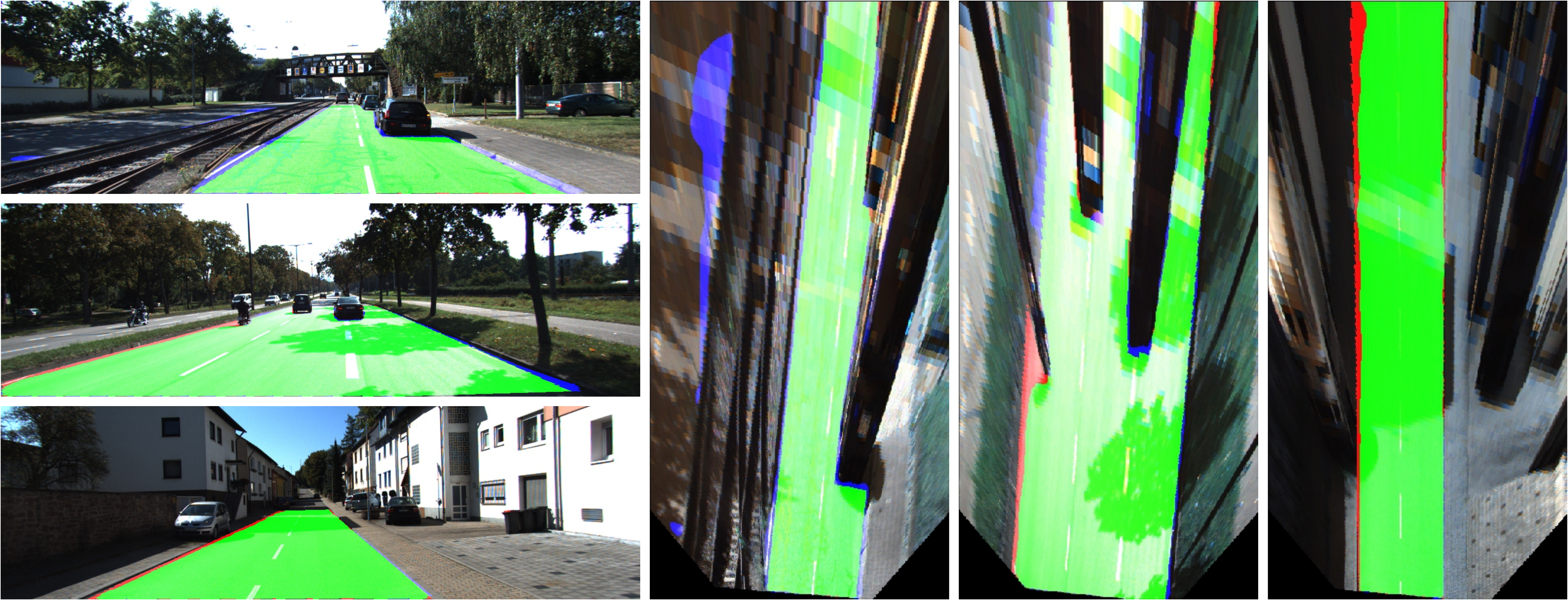}
    \caption{Model C predictions from Kitti-Road evaluation server}
    \label{fig:ExamplesKittiEvaluation}
\end{figure}

Figure \ref{fig:ExamplesKittiEvaluation} presents some examples for road detection on the testing dataset, extracted from the evaluation server results. Here, the road semantic segmentation is overlaid over the camera on the panoramic and BEV perspectives. Green area represents the area of the road that was correctly identified while red indicates false positives and blue corresponds to false positives.

\subsection{Discussion}
\label{sec:Discussion}
Among the state-of-the-art methods on the Kitti-Road dataset that makes use of both camera and LiDAR data, our method ranks TOP-7 according to the Max F-score metric in the general leaderboard. However, TEDNet performs TOP-4 according to the running time and TOP-3 if Average Precision (AP) is taken as the sorting metric.

Methods presented in \cite{Chen2019,Khan2022} show superior performance in both Max F-score and AP. However, these methods perform the Road Detection using a camera-like LiDAR representation of 384$\times$1280 px, that is much larger than our 400$\times$800 px BEV representation, and therefore, impact in the use of memory of these systems. This approach is also followed by \cite{Gu2019Road,Caltagirone2018}. In addition, both methods make use of server-grade equipment, such as Intel Xeon processors, large quantities of RAM memory or Nvidia Titan GPUs. It is undisclosed how the use of lower-performance computers may affect the methods results.

Works such as \cite{Bayón-Gutiérrez2022,Caltagirone2017,Yu2021}, follow a similar data preprocessing approach to the one presented in this work. \cite{Bayón-Gutiérrez2022} corresponds to the authors' previous work in Road Detection, and performs worse than the proposed method in both evaluation metrics while offering better running time thanks to the use of a simpler CNN model. \cite{Caltagirone2017} introduced the use of a BEV representation of the LiDAR data. While offering promising results, the decision to just use LiDAR data affected the method's performance, leading to lower positions on the leaderboard. \cite{Yu2021} presents a BEV representation to the LiDAR data of 200$\times$400 px, similar to the one presented in this work. However, the method performs worse for the AP metric and made use of a high-end CPU and GPU newer than the ones used in this work.

Our method obtains comparable performance with respect to other methods evaluated in the Kitti-Road dataset despite it being deployed in lower performance hardware. This means that, in the case of deploying our model in a real autonomous vehicle, our model would use fewer computational resources, that could be employed for other autonomous driving related systems

\section{Conclusions}
\label{sec:Conclusions}
This article introduced a novel Convolutional Neural Network architecture to perform road surface semantic segmentation in a Bird's Eye View perspective employing RGB camera and LiDAR data.
An initial symmetrical encoder-decoder architecture is first presented and a grid-search technique is applied to find the best hyper-parameter configuration for the model.
An ablation study has been then conducted to exploit different encoder architectures and a total of 6 models have been tested.
K-fold cross-validation was used to compare the ablation study models and declare that, the use of twin non-siamese encoders was beneficial for the model performance.
Predictions from the best model have been submitted to the Kitti-Road benchmark server and can therefore be compared among other state-of-the-art methods.
Despite presenting a much simpler data preprocessing approach and an easy-to-follow CNN architecture, our model offers comparable results to other state-of-the-art methods and it even outperforms most of these in terms of running time and Average Precision.

This work serves as a basis to further investigate in how LiDAR and camera information can be combined to improve road detection. It remains unknown how each of the cited methods would perform in similar circumstances, i.e. in the same hardware, as well as how would the models perform in case of less favorable road conditions, e.g. in low light conditions. Authors continue to work in this research line and expect to further discuss these unknowns in future papers.

\section*{Acknowledgments}
This work is partially supported by Universidad de León, under the "Programa Propio de Investigación de la Universidad de León 2021" grant.

\bibliographystyle{splncs04}
\bibliography{TEDNET}

\end{document}